# A diverse large-scale building dataset and a novel plug-and-play domain generalization method for building extraction

Muying Luo[a], Shunping Ji[a*], Shiqing Wei[a]

[a] *School of Remote Sensing and Information Engineering, Wuhan University, 129 Luoyu Road, Wuhan 430079, China*

* Corresponding author

E-mall addresses: jishunping@whu.edu.cn (S. Ji), luomuying@whu.edu.cn (M. Luo), wei_sq@whu.edu.cn (S. Wei)

---

# Abstract

In this paper, we introduce a new building dataset and propose a novel domain generalization method to facilitate the development of building extraction from high-resolution remote sensing images. The problem with the current building datasets involves that they lack diversity, the quality of the labels is unsatisfactory, and they are hardly used to train a building extraction model with good generalization ability, so as to properly evaluate the real performance of a model in practical scenes. To address these issues, we built a diverse, large-scale, and high-quality building dataset named the WHU-Mix building dataset, which is more practice-oriented. The WHU-Mix building dataset consists of a training/validation set containing 43,727 diverse images collected from all over the world, and a test set containing 8402 images from five other cities on five continents. In addition, to further improve the generalization ability of a building extraction model, we propose a domain generalization method named batch style mixing (BSM), which can be embedded as an efficient plug-and-play module in the frond-end of a building extraction model, providing the model with a progressively larger data distribution to learn data-invariant knowledge. The experiments conducted in this study confirmed the potential of the WHU-Mix building dataset to improve the performance of a building extraction model, resulting in a 6–36% improvement in mIoU, compared to the other existing datasets. The adverse impact of the inaccurate labels in the other datasets can cause about 20% IoU decrease. The experiments also confirmed the high performance of the proposed BSM module in enhancing the generalization ability and robustness of a model, exceeding the baseline model without domain generalization by 13% and the recent domain generalization methods by 4–15% in mIoU.

# 1. Introduction

Automatic building extraction from high-resolution remote sensing images can facilitate a wide range of applications, including topographic map making, population density estimation, land-use management, and urban planning (Clark and Roush, 1984; Freire et al., 2014; Griffiths and Boehm, 2019). The development of deep learning techniques, especially convolutional neural networks (CNNs), has greatly boosted the performance in various image and vision based tasks (He et al., 2016; Szegedy et al., 2015), including building extraction (Liu et al., 2020; Wei et al., 2020; Wei and Ji, 2022; Zhao et al., 2021; Zhou et al., 2022).

Nevertheless, there is still a prominent issue that should be further addressed: the questionable generalization ability of the deep learning models when faced with versatile building structures and architectural styles from all over the world, in addition to the various remote sensing images acquired from different sensors, in diverse weather conditions, different seasons, etc. This problem is recognized to be the biggest challenge currently being faced, and it far surpasses the improvement of structures and parameters in a deep learning based building extraction model.

To handle this generalization problem, large-scale building datasets with various building structures and styles, collected under different acquisition conditions, are urgently needed. On the one hand, the current deep learning based methods are highly data-driven, requiring large-scale and diverse training data to achieve good generalization; on the other hand, the test data should have a certain distribution difference with the training data (e.g., training and test data from different cities), to more properly and practically evaluate the performance of building extraction models. To date, several remote sensing building datasets, such as the Massachusetts buildings dataset (Mnih, 2013), the WHU building dataset (WHU) (Ji et al., 2019), and the Inria aerial image labeling dataset (Maggiori et al., 2017), have been made available openly and freely. However, most of these datasets suffer from several drawbacks. The construction of these datasets, to some extent, caters to the requirement of deep learning methods for the training and test data to share the same or similar data distributions. For example, the training and test data are from the same cities in the Massachusetts buildings dataset and WHU aerial building dataset (Ji et al., 2019), and thus lack diversity. Although the Inria dataset provides a training set and a test set from different cities, the labels of the test set are not openly available. As a result, most studies (Hu and Guo, 2019; Ji et al., 2019; Liu et al., 2020; Zhou et al., 2022) have selected a proportion of the images from each city in the training set for the training, and used the rest for the validation. Nevertheless, the well-performing deep learning models can suffer a severe performance drop with target data which have a quite different distribution to the training data (Chen et al., 2021; Ji et al., 2019; Wittich and Rottensteiner, 2021; Zhang et al., 2021), i.e., out-of-distribution data. Generally speaking, these datasets lack the ability to serve as benchmarks to both train a building extraction model with strong generalization ability and properly evaluate the performance of a model in a more practical scene where the target data are from other cities and are independent of a pre-created training set.

The other major drawback is many datasets, including Massachusetts, Inria, SpaceNet challenge dataset (SpaceNet, 2017), OpenAI dataset (OpenAI, 2018), etc., suffer from some low-quality labels, which greatly impacts the correctness and fairness of evaluating a building extraction model. For example, we discovered that there are about 20% IoU (intersection over union) score gap between using the original Inria and the one repaired by us.

The generalization problem cannot be fully solved by only increasing the quantity of the datasets, because images captured over the broad Earth's surface and under ever-changing atmospheric conditions, various sensors, and seasons form a vast distribution space (Yan et al., 2020; Zhang et al., 2021). This situation has impelled researchers to look for algorithmic solutions. A commonly used methodology is called domain generalization (Gan et al., 2016; Muandet et al., 2013), i.e., learning a model using the limited training data to achieve better generalization for out-of-distribution data. One type of domain generalization methods, the image/data-level methods, which are often called data augmentation methods, attempt to simulate more data from the training data to enlarge the data distribution.

The other type of methods, the feature-level methods, attempt to learn robust and invariant features in a task-specific model (Jin et al., 2020; H. Li et al., 2018) so that the model will be more effective on the unseen data. Nevertheless, most domain generalization methods developed in the computer vision community are designed for the situation where there are several given domains with specific domain labels. For example, a domain of real street scenarios and another of simulated street scenarios are provided in the GTA5 (Richter et al., 2016) and Cityscapes dataset (Cordts et al., 2016), and four domains of photo, art, cartoon, and sketch are provided in the PACS dataset (Li et al., 2017). However, this is rarely the case in remote sensing image processing, where images collected from optical cameras cannot be easily divided into different domains. To date, except for the commonly used data augmentation techniques, such as Gaussian blur and rotation, there is a lack of widely recognized domain generalization baselines, especially for handling complex remote sensing images.

In order to handle the above challenges in generalization, we not only built a large-scale and diverse building dataset collection, but also propose a novel domain generalization method to enhance the generalization ability of deep learning based building extraction models.

The building dataset collection, which we have named the WHU-Mix building dataset, is intended to be adaptable to practical situations, as much as possible. The dataset collection includes a training/validation (trainval) set with diverse building structures and styles from different regions of the globe, and a test set also containing diverse data, but from other global cities. The trainval set contains 43727 512 × 512 image tiles collected from different geographical areas, times, and sensors. The test set contains 8402 512 × 512 image tiles from five cities on five different continents. The WHU-Mix building dataset is larger than the WHU aerial building dataset (Ji et al., 2019), which contains about 8000 tiles of the same size, but is smaller than the CrowdAI dataset (Mohanty, 2018), which has 340000 tiles. Nevertheless, the new dataset features "diversity", which was found to be much more important than quantity in this study.

We also developed a novel, stable, and plug-and-play domain generalization method, named batch style mixing (BSM). The BSM method, as a portable component, dynamically simulates and generates new diverse data to expand the data distribution in a learnable manner during the training of a building extraction model, so that the model can learn more domain-invariant information and perform better on unseen data. In fact, it could be incorporated in any deep learning based building extraction model, because BSM is decoupled from the target model.

The main contributions of this work are summarized as follows:

(1) We introduce an open-source, large-scale, diverse, and highly accurate building dataset, named the WHU-Mix building dataset, which guides remote sensing building extraction research toward more practical applications. When trained with the WHU-Mix building dataset, the same building extraction model obtained an improvement in mIoU score of between 6.43% and 36.52%, compared to the same model trained on the other existing building datasets. In addition, we found and repaired many wrong labels in some of the existing datasets, and proved that these wrong labels can significantly reduce the model performance.

(2) We propose a domain generalization method—BSM—to further improve the generalization and robustness of building extraction models. BSM was developed through a combination of the classic data augmentation methods and a new style mixing method. Although style transfer has recently used in the domain adaptation of paired images in the computer vision and remote sensing communities, its potential in domain generalization has not yet been explored by the two communities. Introducing the style transfer on a batch of training samples with a mixing technology for domain generalization is the brand new idea of this paper. Specifically, BSM expands the data distribution space in the front-end of a deep learning model to handle the complex generalization problem of remote sensing images, and can be easily imbedded in the existing building extraction models. The experiments indicated that the proposed domain generalization method can improve the performance of a deep learning model by 13.17% in mIoU on the diverse test set of the WHU-Mix building dataset.

# 2. Related work

## 2.1 Building extraction datasets of high-resolution remote sensing images

Over the past years, several building extraction datasets have been publicly released by different research groups (see Table 1). These datasets are briefly reviewed as follows:

(1) CrowdAI mapping challenge dataset (Mohanty, 2018). The CrowdAI dataset is a very large scale building extraction dataset consisting of 341058 300 × 300 images collected from Google Earth. The annotation accuracy of this dataset is high. However, we found that the images in this dataset tend to be acquired in the same local region, and thus lack diversity.

(2) Massachusetts buildings dataset (Mnih, 2013). The Massachusetts buildings dataset is an early building extraction dataset consisting of 804 1500 × 1500 aerial images acquired in the Boston area of the U.S. The images have a relatively low resolution of 1 m, and there are some obvious annotation errors.

(3) OpenAI dataset (OpenAI, 2018). The OpenAI dataset was released as part of the 2018 OpenAI Tanzania Building Footprint Segmentation Challenge Competition. The images were collected in Tanzania by unmanned aerial vehicle (UAV), with a very high resolution of about 0.07 m. However, there are a large number of errors in the annotations.

(4) WHU building dataset (Ji et al., 2019). The WHU building dataset is a multi-source building extraction dataset providing high-quality labels, which consists of three subsets. Among these subsets, the aerial imagery subset is the most widely used for building extraction in the remote sensing community.

(5) SpaceNet challenge dataset (SpaceNet, 2017). The SpaceNet challenge dataset was designed for the task of building extraction in satellite images. The dataset consists of images collected from five cities on five continents. Unfortunately, the annotation accuracy is low, and the dataset cannot be used to properly evaluate the performance of a building extraction model.

(6) Learning Aerial Image Segmentation From Online Maps (LAIS) dataset (Kaiser et al., 2017). The LAIS dataset was designed to evaluate the performance of deep learning based models when trained with noisy data. Three land-cover classes are annotated: building, street, and background. This dataset contains images of four cities, which were downloaded from Google Maps, and their corresponding relatively inaccurate annotations obtained from OpenStreetMap (OSM), with a displacement as large as 10 pixels. This dataset also provides relabeled Potsdam images from the ISPRS 2D Semantic Labeling Contest dataset (Gerke et al., 2014).

(7) Inria aerial image labeling dataset (Maggiori et al., 2017). The Inria dataset was released to benchmark the generalization ability of different building extraction methods, for a similar purpose as this paper. It contains a training set and a test set, and the images of the two sets are from different cities. However, as the annotations of the test set are not openly available from the provider, most studies have divided the training set for the training and testing (Hu and Guo, 2019; Zhou et al., 2022), resulting in the dataset having a limited ability to properly evaluate a model's generalization capability. Moreover, there are many obvious annotation errors in the dataset, which have an adverse impact on the training and testing.

As a summary, although these datasets have greatly facilitated the development of building extraction study, the lack of diversity and the low-quality labels are the main concerns of the existing datasets.

Table 1 Seven publicly available datasets and the WHU-Mix building dataset built in this study. '~' denotes "about".

| Dataset | Subsets | Number of image tiles | Image size | Image spatial resolution (m) | Area ($km^2$) | Image type |
|---|---|---|---|---|---|---|
| CrowdAI mapping challenge dataset (2018) | / | 341058 | 300 × 300 | unknown | / | satellite |
| Massachusetts buildings dataset (2013) | / | 151 | 1500 × 1500 | 1 | 340 | aerial |
| Open AI dataset (2018) | / | 13 | ~40000 × 40000 | ~0.07 | 102 | aerial |
| WHU building dataset (2018) | Aerial imagery dataset | 8188 | 512 × 512 | 0.3 | 193 | aerial |
| | Satellite dataset I (global cities) | 204 | | 0.3–2.5 | ~5 | satellite |
| | Satellite dataset II (East Asia) | 17388 | | 0.35 | 558 | satellite |
| SpaceNet challenge dataset (2017) | Rio de Janeiro | 6940 | 438 × 406 | 0.5 | 308 | satellite |
| | Vegas | 3851 | 650 × 650 | 0.3 | 146 | |
| | Paris | 1148 | | | 44 | |
| | Shanghai | 4582 | | | 174 | |
| | Khartoum | 1012 | | | 38 | |
| Learning Aerial Image Segmentation From Online Maps dataset (2017) | Berlin | 200 | ~3000 × 3000 | ~0.1 | 10 | aerial |
| | Chicago | 497 | | | 51 | |
| | Paris | 625 | | | 60 | |
| | Potsdam | 24 | | | 2 | |
| | Zurich | 375 | | | 36 | |
| Inria aerial image labeling dataset (2017) | Training set (Austin, Chicago, Kitsap County, Western Tyrol, Vienna) | 180 | 5000 × 5000 | 0.3 | 405 | aerial |
| WHU-Mix building dataset | Trainval set | 43727 | 512 × 512 | 0.091–2.5 | 1080 | aerial/ satellite |
| | Test set | 7718 | | | 133 | |

## 2.2 Domain generalization

The domain shift problem poses challenges for cross-domain recognition tasks, e.g., the land-cover classification of multi-source or multi-temporal remote sensing images, leading to a significant performance drop for the data-driven deep learning models. To narrow the discrepancy across domains, there are two main approaches. Domain adaptation (Tuia et al., 2016) involves using both the labeled training data in the source domain and the unlabeled test data in the target domain in the training process, to learn a target-adapted model. Differing from domain adaptation, domain generalization (Zhou et al., 2021a) is independent of target domains. It attempts to make the model learn domain-invariant knowledge from only the labeled source domain, and is performed directly on an arbitrary target domain that has not been seen before. Therefore, compared to domain adaptation, domain generalization is more challenging because no knowledge from the target domain can be accessed. However, domain generalization has more practical value because there is no need to repeat the training process of a model every time for different target domains, especially when the source domain is an extraordinarily large dataset (e.g., ImageNet (Deng et al., 2009)).

As an image/data-level method, data augmentation is one of the important categories of domain generalization (Zhou et al., 2021a). Data augmentation involves manipulating the training data to generate diverse samples, to help the model learn domain-invariant knowledge. The classic data augmentation operation is the simplest, cheapest, and most common way to prevent overfitting and improve the generalization ability of a deep learning model. Geometric data augmentation usually includes flipping, rotation, scaling, cropping, and so on. Color augmentation usually consists of color jitter and adding random noise, to enhance the robustness of a model to image color variations.

However, the classic data augmentation operations are fundamentally linear transformations restricted by the range of the transformation parameters, leading to a limited ability to augment images. Some recent augmentation methods introduce a style transfer technique to generate more diverse and rich data. Nevertheless, most style transfer based methods require auxiliary datasets as the augmentation reference, or require known different source domain categories within the dataset to be the references for each other. For example, the domain randomization and pyramid consistency (DRPC) method (Yue et al., 2019) applies a generative adversarial network (GAN)-based style transfer network named CycleGAN (Zhu et al., 2017) to synthesize images, randomly transferring the styles of real images to have similar visual appearances to ImageNet classes. Borlino et al. (2020) applied an off-the-shelf feedforward stylization based method known as Adaptive Instance Normalization (AdaIN) (Huang and Belongie, 2017) to transfer styles between different domains. Mixup (Zhang et al., 2018) involves performing linear interpolation between any two image pairs and their labels to generate a continuous data distribution. However, Mixup cannot be directly applied to the pixel-level semantic segmentation task because it can result in semantic confusion. Inspired by Mixup, Mixstyle (Zhou et al., 2021b) mixes deep feature statistics of the training images to synthesize new domains.

In contrast to data augmentation, in which the model is expected to learn from the augmented diverse data, domain generalization at the feature level attempts to provide the domain-invariant deep features of the training data for a task model. Therefore, the feature-level domain generalization must be combined with a task model. Some earlier studies attempted to align the features of the source domains by explicitly minimizing the distance function between the source domain feature distributions. Examples of such methods are maximum mean discrepancy (H. Li et al., 2018), first/second order correlation (Sun et al., 2016), and the linear-dependency domain generalization (LDDG) method (H. Li et al., 2020), which uses Kullback-Leibler divergence as a linear-dependency regularization term. Other studies have adopted the adversarial learning technique (Goodfellow et al., 2014) to align the features of the source domains implicitly. For example, Y. Li et al. (2018) proposed a conditional invariant adversarial network for domain-invariant representation learning. Matsuura and Harada (2020) proposed a method named the mixture of multiple latent domains (MMLD), which performs adversarial training using data of latent domains divided by iterative clustering. Recently, the feature normalization technique has been introduced to learn domain-invariant features. IBN-NET (Pan et al., 2018) combines instance normalization and batch normalization to enhance the generalization and learning ability of a deep learning model. Following the work of IBN-NET, the RobustNet method (Choi et al., 2021) introduces an instance selective whitening loss, which separates the domain-specific and domain-invariant features by comparing the features of the images before and after color augmentation (i.e., photometric transformation), and selectively suppresses the domain-specific features. However, as the feature-level domain generalization methods must be coupled with a specific task model, they lack flexibility. Furthermore, as the learned features are in fact only invariant to the source domains, their reliability in unseen domains is not fully guaranteed, and will likely result in an unstable performance.

In addition, some general machine learning strategies have also been explored for some parts of domain generalization. The main strategies include meta-learning (D. Li et al., 2018), ensemble learning (Mancini et al., 2018), and self-supervised learning (Carlucci et al., 2019).

It should be noted that most of the above domain generalization methods (Borlino et al., 2020; H. Li et al., 2018; Y. Li et al., 2018; Sun et al., 2016; Yue et al., 2019) developed in the computer vision community can only be applied when there are several defined domains in the training data. Therefore, these methods cannot be implemented for most remote sensing image datasets without domain classification. There is therefore an urgent need to develop a new domain generalization baseline that exceeds the conventional geometric and color augmentation for remote sensing image processing, including building extraction.

# 3. WHU-Mix building dataset

Although several building extraction datasets have been publicly released, as reviewed in Section 2.1, and have contributed to the development of building extraction methods, the diversity they represent, including the building styles and geographic distributions, is not perfect, and the quality of some datasets is deeply affected by non-ignorable label errors, both of which have hindered the development of further studies of building extraction. To alleviate this problem, we built the WHU-Mix building dataset, which is an open-source, diverse, large-scale, and high-quality dataset that is partially based on the existing datasets. This was done to better simulate the situation of practical building extraction, to measure more reasonably the real performance of a deep learning model, and to evaluate more conveniently the generalization ability of a model on different remote sensing images acquired from different sources and places.

## 3.1 Overview of the WHU-Mix building dataset

As listed in Table 2, the WHU-Mix building dataset consists of two parts: a training/validation (trainval) set and a test set. The trainval set is a collection of data from two datasets, i.e., the WHU building dataset (Ji et al., 2019), and the Inria dataset (Maggiori et al., 2017) which we newly edited, so as to integrate and fully utilize the existing works, and a great many newly acquired samples, to further enrich the diversity. The trainval set consists of 43727 512 × 512 image tiles collected from cities and countryside all over the world and from various remote sensing platforms. We randomly selected 39346 remote sensing images (i.e., 90% of the trainval set) from the trainval set as the training set, and the 4381 remaining images formed the validation set.

For the evaluation of the generalization of building extraction models trained with the trainval set, the test set we prepared is composed of 8402 512 × 512 image tiles, as well as the original uncropped remote sensing images, which were obtained from five cities on five continents. Each subset of the test set is named with the "country_city" rule, as shown in Table 2. In the test set, the data of US_Kitsap, DE_Potsdam, and SD_Khartoum are newly edited from existing datasets, and the data of CN_Wuxi and NZ_Dunedin are newly acquired.

There is no geographic overlap between the training and test sets in the globally distributed WHU-Mix building dataset, making the building extraction task very challenging.

Table 2 Components of the WHU-Mix building dataset, including the trainval set consisting of data from the WHU building dataset (Ji et al., 2019), a subset of the Inria dataset (Maggiori et al., 2017), and new samples, and a test set consisting of a five-city dataset from five continents.

| Division | Dataset | Number of image tiles | Image spatial resolution (m) |
|---|---|---|---|
| Trainval set | WHU building dataset | 12430 | 0.3–2.5 |
| | Subset of edited Inria (Western Tyrol, Vienna) | 5832 | 0.3 |
| | New samples | 25465 | 0.15–0.64 |
| Test set | US_Kitsap | 3600 | 0.3 |
| | CN_Wuxi | 1409 | ~0.17 |
| | NZ_Dunedin | 1312 | 0.2 |
| | DE_Potsdam | 1176 | 0.09 |
| | SD_Khartoum | 905 | 0.3 |

Compared to the existing datasets listed in Table 1, the WHU-Mix building dataset has the following notable properties.

(1) Large scale, wide distribution, and diversity

The WHU-Mix building dataset consists of 51445 512 × 512 image tiles with 0.09–2.5 m spatial resolutions, captured by a wide range of aerial and satellite sensors from about 58 cities worldwide, as shown in Fig. 1, covering an area of about 1213 km$^2$. It is much larger than most of the other datasets listed in Table 1. The different sources and imaging conditions, different spatial resolutions, and different buildings styles of the local regions all contribute to the diversity of the sample distribution (see Fig. 2) of the dataset, which is crucial to improve the generalization ability of deep learning based models (K. Li et al., 2020). The WHU-Mix building dataset is smaller than the CrowdAI dataset (Mohanty, 2018); however, in this paper, we demonstrate that the newly built dataset significantly surpasses the CrowdAI dataset, and prove that diversity is more important than quantity.

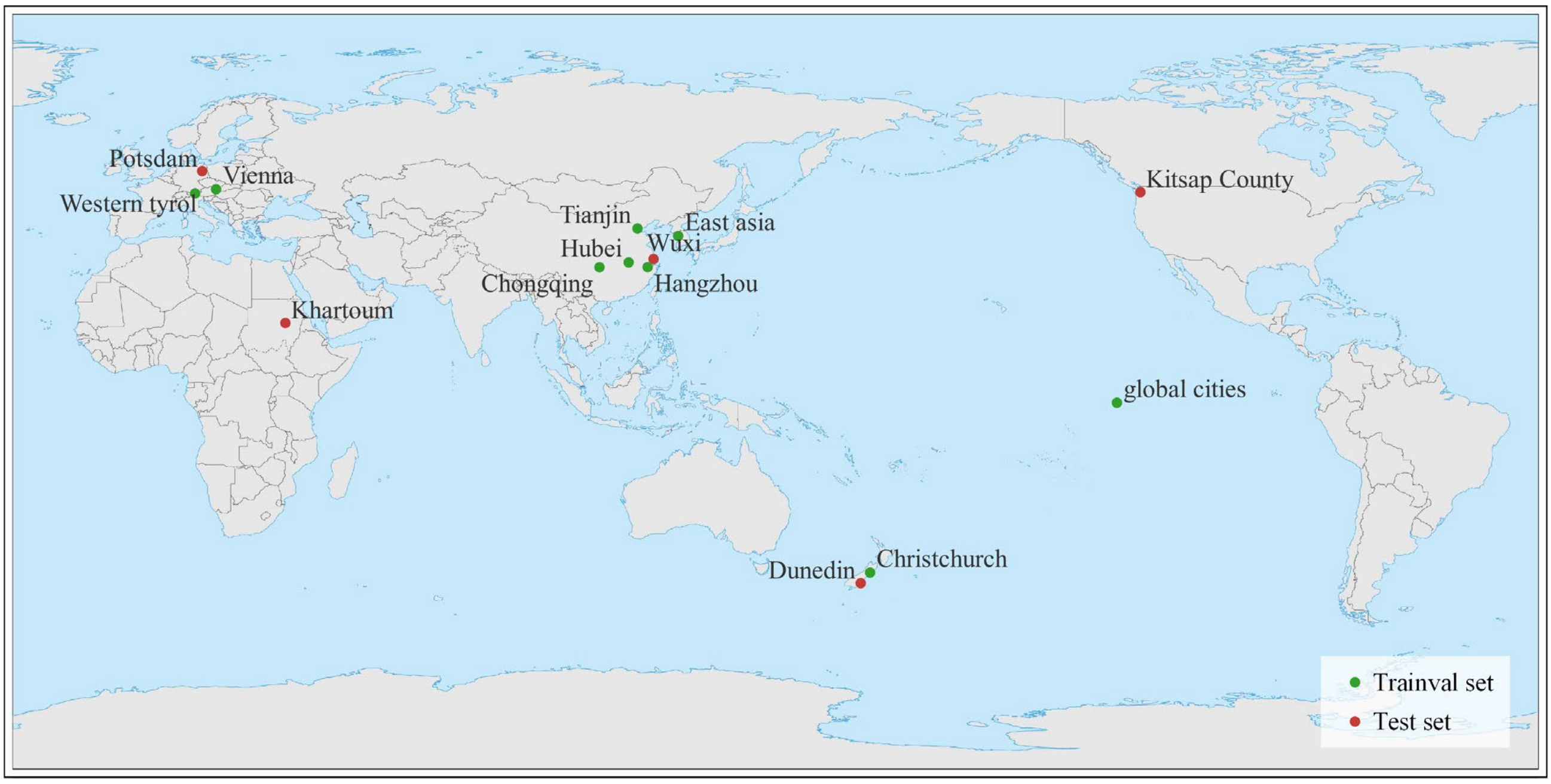

Fig. 1. The geographic distribution of the images of the WHU-Mix building dataset. The green dots and red dots represent images of the trainval set and test set, respectively. The global cities in the original WHU building dataset consist of images from about 45 cities.

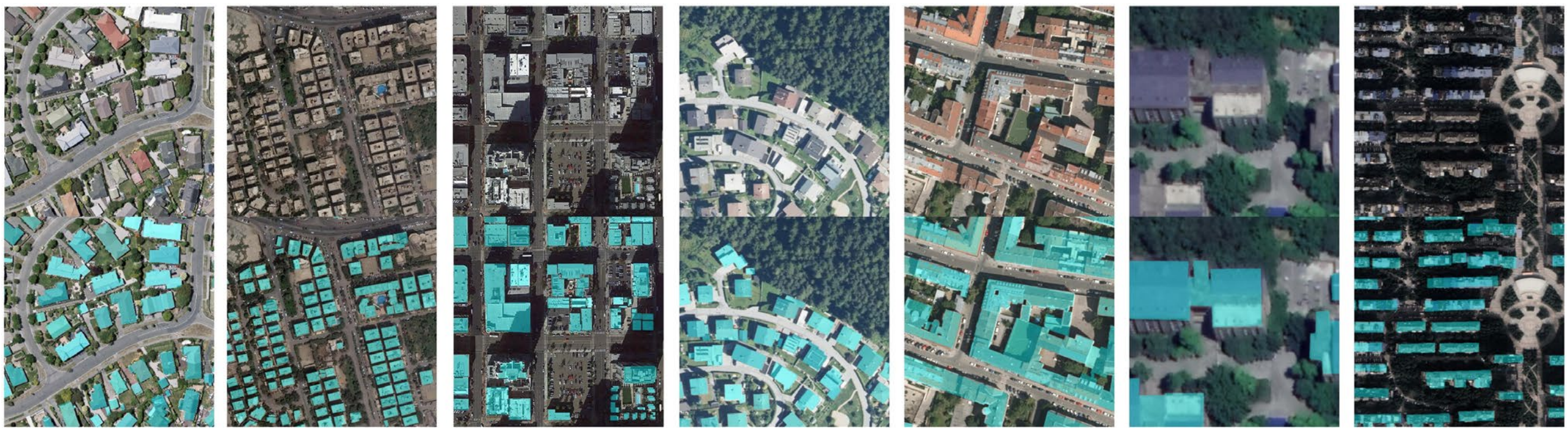

(a) Christchurch (b) Cairo (c) Santiago (d) Western Tyrol (e) Vienna (f) Chongqing (g) Hubei

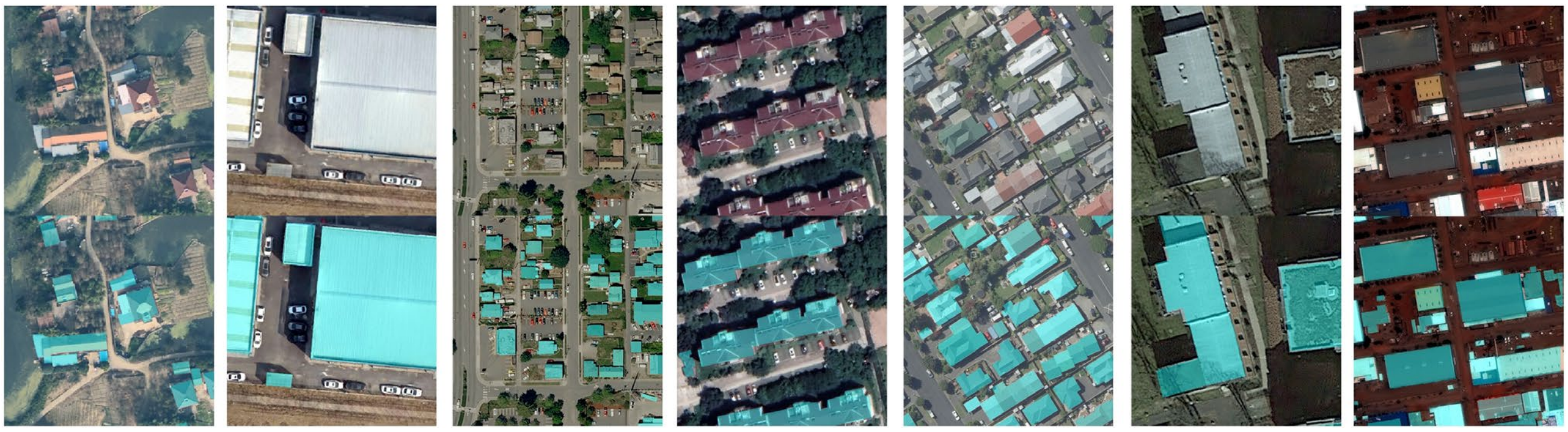

(h) Hangzhou (i) Tianjin (j) Kitsap County (k) Wuxi (l) Dunedin (m) Potsdam (n) Khartoum

Fig. 2. Examples of various images and the corresponding labels (cyan mask) in the WHU-Mix building dataset.

(2) High accuracy and uniform format.

The WHU-Mix building dataset provides high-accuracy labels. For the labeled images from the existing datasets, we checked the labels and corrected the labeling errors. For the newly acquired images without labels, we labeled them manually and then double-checked the labeling. The WHU-Mix building dataset also provides a uniform format for the data collected from different sources with different formats. The image tiles and corresponding building label maps are stored in TIFF format. The building label maps are Boolean maps, where the building pixels and backgrounds are denoted as 255 and 0, respectively. The images and labels of the trainval and test sets are stored in corresponding folders and named with the uniform "city + number" rule.

## 3.2 Construction of the WHU-Mix building dataset

Although we collected some of the samples from the existing datasets, the whole construction process for the WHU-Mix building dataset took more than six months. The WHU building dataset (Ji et al., 2019) is directly included in the WHU-Mix building dataset. Some parts of three other datasets, i.e., Inria (Maggiori et al., 2017), SpaceNet (SpaceNet, 2017), and LAIS (Kaiser et al., 2017), which feature some erroneous annotations, were edited and corrected. Finally, new remote sensing data were collected and manually labeled.

### 3.2.1 Collection of the existing datasets

The WHU building dataset (Ji et al., 2019) is a widely used dataset with high-accuracy building maps. We incorporated the WHU building dataset as a part of the trainval set of the WHU-Mix building dataset. Specifically, for the images from the aerial imagery dataset and satellite dataset I (global cities), we included them in the proposed dataset directly. For satellite dataset II (East Asia), as the images were captured in an area with sparse buildings, 13350 out of the 17388 512 × 512 image tiles have only backgrounds. We therefore removed them and kept the remaining 4038 images.

### 3.2.2 Revision of the existing datasets

We revised the labels for the Kitsap County, Western Tyrol, and Vienna images from the Inria dataset (Maggiori et al., 2017), the Khartoum images from the SpaceNet dataset (SpaceNet, 2017), and the Potsdam images from the LAIS dataset (Kaiser et al., 2017). There are a great many obvious errors in the building labels of these datasets, including non-existing, missing, displaced, and inaccurate delineation (Fig. 3). To construct a dataset with high-quality labels, we first converted the different label formats of all the datasets into vector format. We then manually corrected the outlines of all the building samples using ArcGIS 10.2. After the revision, we converted the edited vector maps back to the uniform raster format. Finally, we added the data of Western Tyrol and Vienna into the trainval set of the WHU-Mix building dataset, and included the data of Kitsap County, Khartoum, and Potsdam in the test set.

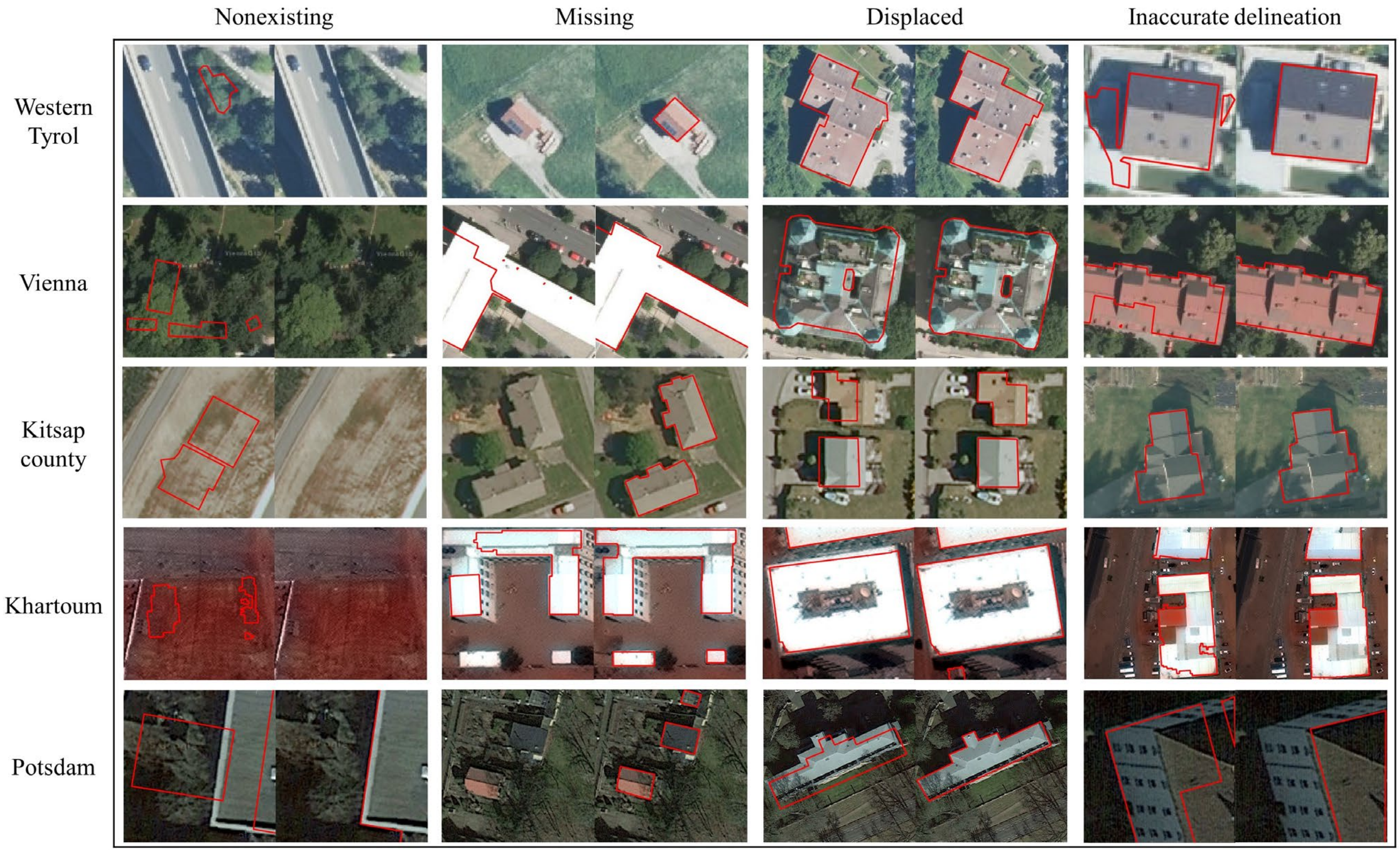

Fig. 3. Example images of the original datasets and the revised images. The left of each column shows the original labels, and the right shows the edited labels.

### 3.2.3 New samples

A large amount of new samples were collected to enrich the trainval set. The samples were made up of 25465 512 × 512 diverse image tiles acquired in the cities of Chongqing, Tianjin, and Hangzhou, and Hubei province in western, northern, eastern, and central China, respectively, from various remote sensing resources. Among the different images, the images of Chongqing have a resolution of 0.15 m and a total coverage area of 32 $km^2$. The buildings in Chongqing are mostly built on the mountainsides, with varied shapes. The images of Tianjin cover a 3 $km^2$ area with a 0.1 m ground resolution, and the building types are dominated by large factories and residential areas. The images of Hangzhou have a ground resolution of 0.2 m and cover an area of 156 $km^2$ in the rural area, where the buildings are sparsely distributed, and water bodies and farmland are the main land-cover types. The images of Hubei province have a resolution of 0.64 m and cover a total area of 427 $km^2$ in several rural and urban regions of Hubei province. The buildings in both the rural and urban regions are densely distributed; however, the buildings in the rural regions are basically low-rise buildings, while the buildings in the urban regions are typically multi-story buildings.

We also collected data from two cities—Wuxi in China and Dunedin in New Zealand—for inclusion in the test set, so that the test set contains images of five typical cities from five continents. The data of Wuxi (named CN_Wuxi in Table 2) were collected from Google Earth, and the data of Dunedin (named NZ_Dunedin in Table 2) were acquired from the Land Information service of New Zealand (LINZ, 2022).

For the above newly acquired remote sensing data, we manually delineated the vector building maps in ArcGIS 10.2. To ensure the label quality and avoid misjudgments, we relied on one additional annotator for a complete check, and three annotators for random checks of a proportion of around 20%, respectively. Finally, we converted the vector building maps to raster maps. Examples of the new samples in the WHU-Mix building dataset are shown in Fig. 2 (f–i, k, l).

# 4. Methodology

In addition to providing a diverse building dataset, the other main contribution of this study is that we propose a simple but very effective and efficient plug-and-play domain generalization method for boosting the performance of a modern deep learning based building extraction model on various target remote sensing images. The proposed domain generalization method uses only the available training datasets to enable the building extraction model to learn data-invariant features. That is to say, it is not necessary to train a new model every time for different target images, as is required with the domain adaptation methods. The pretrained model obtained by the proposed method can be directly applied to target images captured from different sensors, times, regions, and imaging conditions. The basic idea of the proposed method is introduced in Section 4.1, and the implementation details are described in Section 4.2.

## 4.1 The proposed batch style mixing domain generalization method

The proposed image-level domain generalization method, which we call batch style mixing (BSM), can be easily plugged into the front-end of a deep learning based building extraction model. BSM draws inspiration from the idea of style transfer for domain adaptation (Choi et al., 2019; Kim and Byun, 2020). However, in the domain generalization problem, style transfer is used to generate diverse new images and expand the sample distribution space, instead of aligning the styles of the source images in the training set and the target images to be processed in a domain adaptation problem. Specifically, we propose generating diverse new samples in an ever-changing batch of the training data in the training process, where, the images in the batch are first processed by random data augmentation, and their styles are then mixed (and thus expanded) by the idea of mixing (Zhang et al., 2018). The workflow of BSM is shown in Fig. 4, which includes three main steps: geometric and color augmentation, shuffling, and a style transfer network for obtaining the mixed-style new samples.

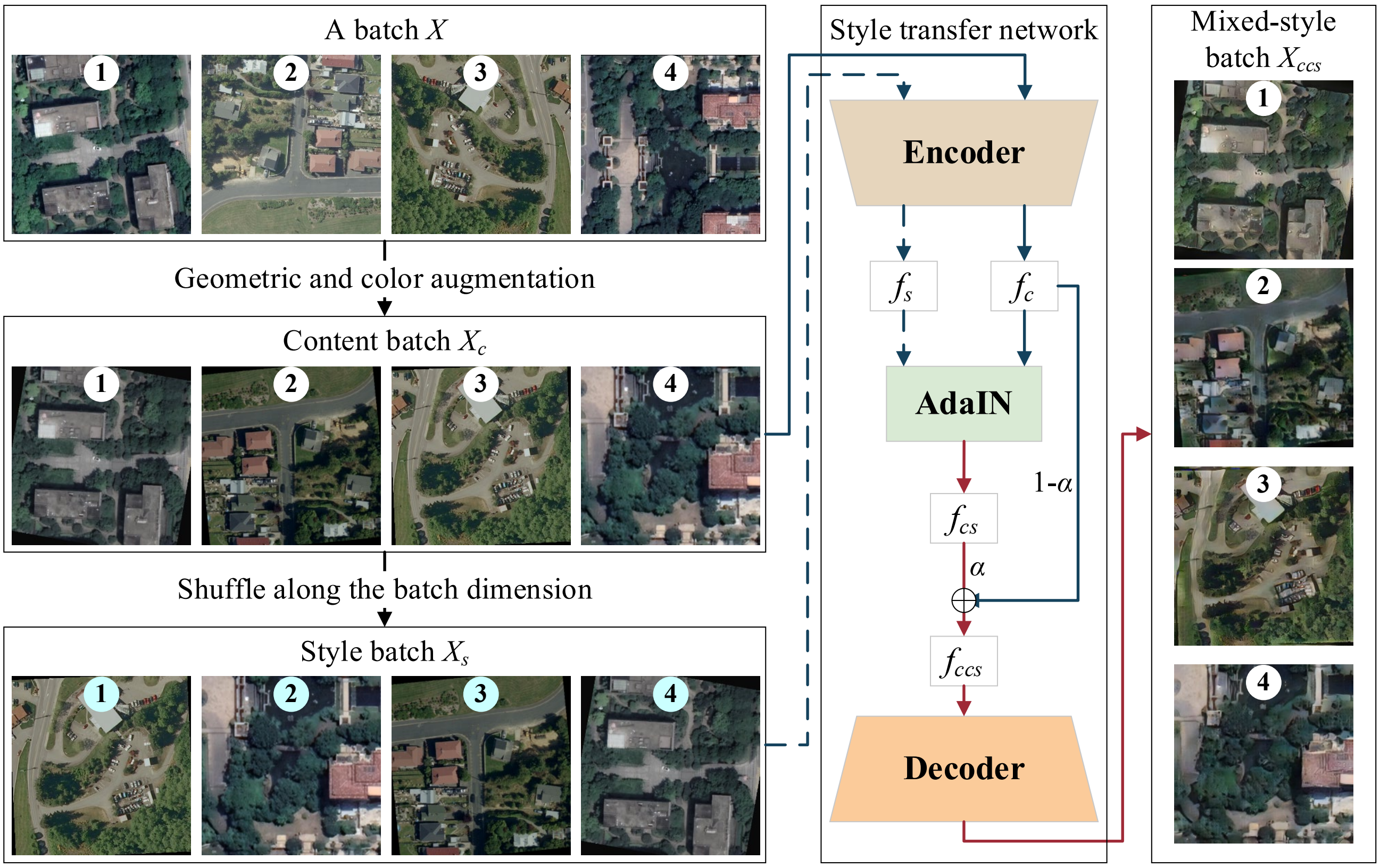

Fig. 4. An overview of batch style mixing.

As shown in Fig. 4, for a batch $X$ with a given batch size (here 4), some common geometric and color augmentations are performed on the images in $X$ to obtain a *content* batch $X_c$. The images in $X_c$ are then shuffled along the batch dimension to obtain a *style* batch $X_s$. The shuffling operation prepares pairs of images (with the same order numbers), for which the styles will next be mixed with each other. The content and style batches $X_c$ and $X_s$ are both fed into the style transfer network. $f_c$ and $f_s$ are the feature outputs from the encoder of the style transfer network corresponding to $X_c$ and $X_s$. We choose AdaIN (Huang and Belongie, 2017) to process $f_c$ and $f_s$, i.e., to re-normalize $f_c$ by replacing its channel-wise mean and standard deviation with that of $f_s$. In fact, AdaIN operates in the same way as the Wallis filter (Li et al., 2006), but on the feature maps. As a result, the style information represented by feature $f_{cs}$ is the same or similar to $f_s$, but the content information remains unchanged. $f_{cs}$ can be denoted as:

$$f_{cs} = \sigma(f_s)\left(\frac{f_c - \mu(f_c)}{\sigma(f_c)}\right) + \mu(f_s) \quad (1)$$

where $\sigma(*)$ and $\mu(*)$ are the channel-wise mean and standard deviation of the features. Then, as linear interpolation has been shown to be an effective way to mix samples (Zhang et al., 2018), we also perform linear interpolation between the feature maps $f_c$ and $f_{cs}$ for the mixing styles. Feature $f_{ccs}$ is then obtained, as shown in Eq. (2):

$$f_{ccs} = \alpha f_{cs} + (1-\alpha) f_c \quad (2)$$

where the trade-off parameter $\alpha$ determines the degree of mixing of the styles, which was set to 0.5 in this study.

Finally, the feature $f_{ccs}$ is fed into the decoder of the style transfer network to output a mixed-style batch $X_{ccs}$. The $i$-th image in $X_{ccs}$ has the same semantic content as the $i$-th image in $X_c$, and its style is a mixture of the $i$-th images in $X_c$ and $X_s$. The images in a batch are constantly changing during the training, which ensures that the combination of data augmentation and mixing creates a huge distribution space.

## 4.2 The implementation of BSM for building extraction

### 4.2.1 Building extraction framework with BSM

The overall building extraction framework with the proposed domain generalization method is shown in Fig. 5. The framework consists of two parts: the front-end BSM for domain generalization, as described in Fig. 4, and a back-end building extraction model, which can be an arbitrary semantic segmentation model. In this study, we applied a modified model named AT-MAFCN, which is described in detail in Section 4.2.2.

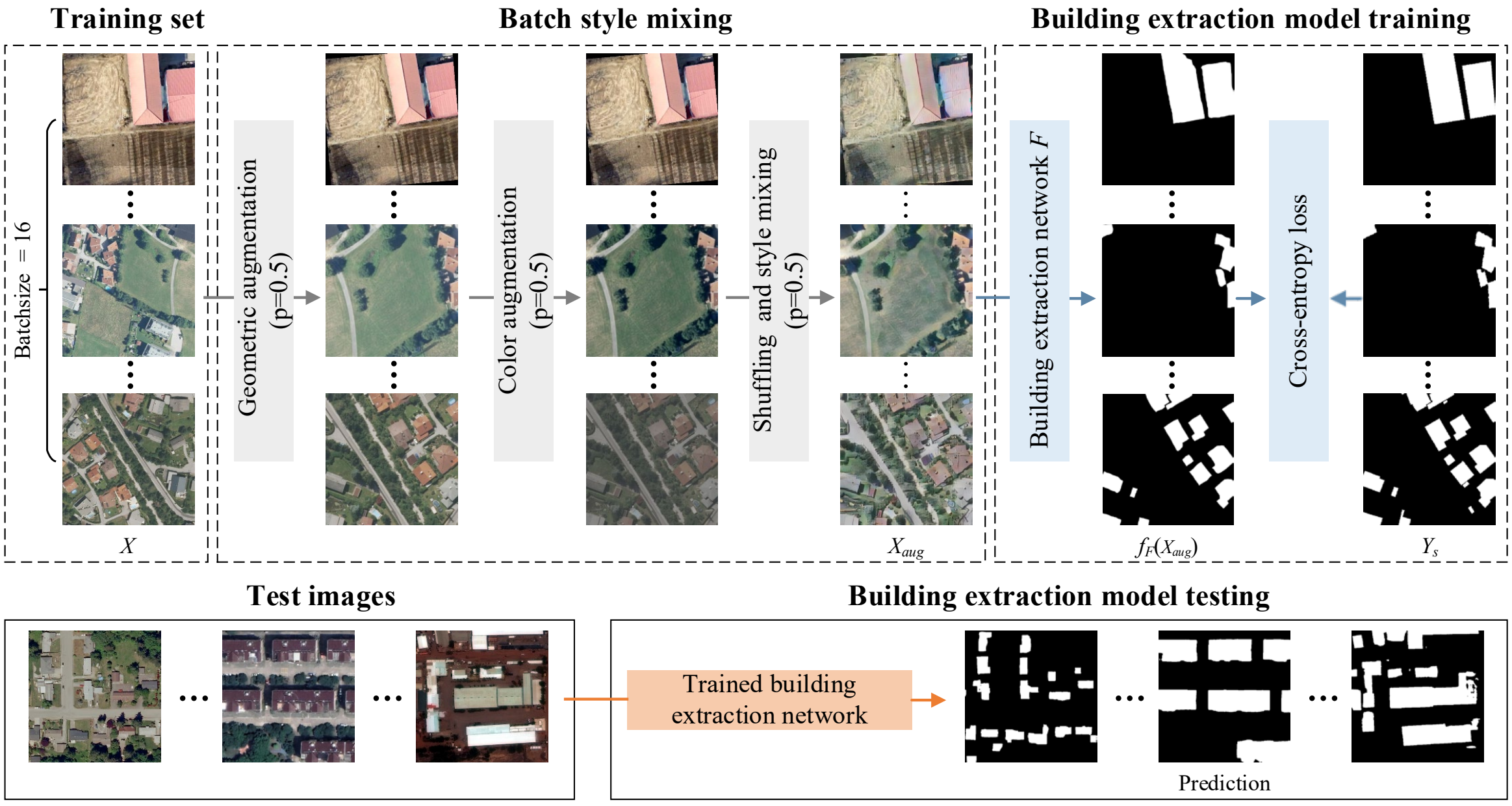

Fig. 5. The implementation procedure. P = 0.5 means that the module is performed with a probability of 0.5.

The training process is shown in the dashed box in Fig. 5. During the training process, the training data are first fed into the BSM module. Specifically, in each iteration, a batch $X$ (with batch size 16) is sequentially forwarded from the training set to the geometric augmentation, color augmentation, and style mixing, to obtain an augmented batch $X_{aug}$. The operations in the three submodules of the BSM module are each activated with a probability of 0.5. For the geometric augmentation, random horizontal flip, random vertical flip, random rotation within 30 degrees, random scaling of the image size within 0.5–2.0, and cropping (or filling) to the original size are applied. The color augmentation includes the adjustment of the brightness, color balance, contrast and sharpness, and random Gaussian blur.

The augmented images in batch $X_{aug}$ are then input into the building extraction model $F$ to complete the standard training process and obtain the prediction maps $f_F(X_{aug})$. As the number of iterations increases, BSM provides a progressively larger and larger data distribution for the building extraction model to learn data-invariant features. Cross-entropy loss is used to calculate the pixel-level classification loss between $f_F(X_{aug})$ and the corresponding label $Y_s$. For each image $x_{aug}$ in $X_{aug}$, the loss is calculated as follows:

$$L_{ce}(x_{aug}) = -\frac{1}{HW}\sum_{h=1}^{H}\sum_{w=1}^{W}\sum_{c=1}^{C}\left[ y_s^{(h,w,c)} \cdot \log(f_F(x_{aug}))^{i,j,c} \right] \tag{3}$$

where $y_s$ is the corresponding label of $x_{aug}$. $H$ and $W$ are the height and width of the prediction map $f_F(x_{aug})$, and C is the number of categories (for building extraction, C is 2).

The building extraction model trained along with BSM thus has a strong generalization capability for images that do not match the distribution of the training set. The test process is depicted in the solid box in Fig. 5. The various test images are fed into the pretrained building extraction model without BSM, to output the prediction maps.

### 4.2.2 The implemented building extraction model

The implemented building extraction model—AT-MAFCN—is modified from a multi-scale semantic segmentation network known as MA-FCN (Wei et al., 2020). The structure of AT-MAFCN is illustrated in Fig. 6. We employ the same VGG-16 (Simonyan and Zisserman, 2015) encoder as MA-FCN. There are three main differences between AT-MAFCN and MA-FCN.

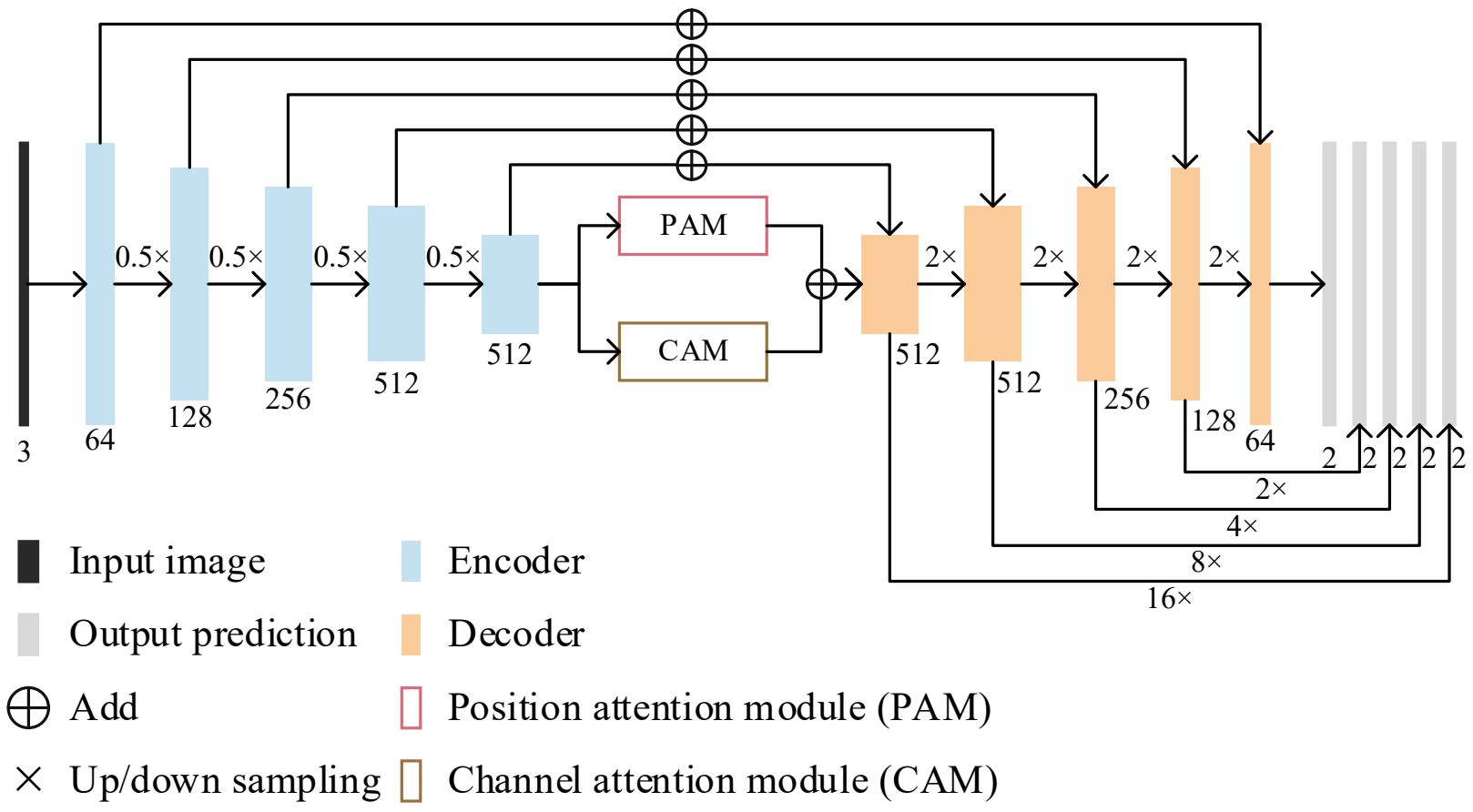

Fig. 6. Structure of AT-MAFCN.

Firstly, two parallel attention modules are placed between the last layer of the encoder and the first layer of the decoder, one of which is the channel attention module (CAM) and the other is the position attention module (PAM) (Woo et al., 2018), to enhance the representation ability of the deep fully convolutional network (FCN) structure.

Secondly, we believe that learning from the output feature maps at a wider scale can minimize the impact of different sizes of buildings and various resolutions of images. Therefore, the output feature maps of all five scales in the decoder are upsampled, instead of the last four scales, as in the original MA-FCN. The pixel-level classification loss is then calculated using the output feature maps obtained from scales (1/16, 1/8, 1/4, 1/2, and 1/1) with corresponding weights {0.25, 0.25, 0.25, 0.25, 0.5} in the training process. Eq. (4) is the final loss:

$$L_{ce}(x_{aug}) = 0.25\sum_{i=1}^{4} L_i + 0.5L_5 \tag{4}$$

where $L_i$ ($i$ = 1, 2, 3, 4, 5) represent the losses of the five prediction maps at scales (1/16, 1/8, 1/4, 1/2 and 1/1), respectively. $L_i$ is implemented as shown in Eq. (3).

Thirdly, in the test process, the prediction map is only output from the feature of the last layer (scale 1/1), instead of the concatenated feature maps of five layers in the decoder. This is because, after the multi-scale training, the prediction map from the last layer is what we really want, and maps with other scales are not necessary in the testing, and the prediction efficiency is increased.

# 5. Experimental results and analysis

The training settings and the evaluation metrics used in the experiments are described in Section 5.1. The experiments conducted in this study to evaluate the WHU-Mix building dataset are reported in Section 5.2. Finally, the experimental evaluation of the proposed domain generalization method is provided in Section 5.3.

## 5.1 Experimental setup

### 5.1.1 Training settings

In all the experiments, we used a single NVIDIA RTX A6000 48 GB GPU and an Intel Xeon Silver 4109T CPU. Among the building extraction models used in the experiments, AT-MAFCN and UNet (Ronneberger et al., 2015) employed the VGG-16 (Simonyan and Zisserman, 2015) backbone pretrained on the ImageNet dataset (Deng et al., 2009) in the training process, and DeepLab V3+ (Chen et al., 2018) employed the ResNet-101 (He et al., 2016) backbone pretrained on the ImageNet dataset. The structures and weights of the encoder and decoder of the style transfer network (as shown in Fig. 4) were the same as those in Huang and Belongie (2017). The building extraction models were trained by the Adam optimizer (Kingma and Ba, 2015), with a weight decay of $5 \times 10^{-5}$ and an initial learning rate of $1 \times 10^{-4}$. The learning rate was dynamically adjusted by a poly learning strategy with decay parameter of 0.9. To fit the GPU capacity, the batch size was set to 16, and the images used in the experiments were uniformly cropped into the size of 512 × 512 pixels. During the test process, the images were fed into the trained model with a batch size of 1.

### 5.1.2 Evaluation metrics

The test set of the WHU-Mix building dataset was used to assess the building extraction accuracy. The intersection over union (IoU) for the test data of the five cities was computed. The IoU score is the ratio between the intersection and union of the true positive pixels and the building pixels predicted by the model, which can be defined as follows:

$$\mathrm{IoU} = \frac{\mathrm{TP}}{\mathrm{TP{+}FP{+}FN}} \tag{5}$$

where TP, FP, and FN are the number of true positive, false positive and false negative pixels, respectively.

As a global metric, the mean intersection over union (mIoU) over all the images in the five cities of the test set is reported. The values of IoU and mIoU are in the range of 0 to 1 (they are reported as percentages (%) in the experiments). The closer the values are to 1, the better the building extraction results. In addition, as an auxiliary indicator, the training times of the building extraction models using the different datasets and domain generalization methods are also reported.

## 5.2 Evaluation of the WHU-Mix building dataset

The experiment described in Section 5.2.1 compared the performance of AT-MAFCN, pretrained on different building extraction datasets, on the test set of the WHU-Mix building dataset, to evaluate the impact of the size and diversity of the dataset. The experiment described in Section 5.2.2 was dedicated to the evaluation of the influence of data quality on the building extraction performance.

### 5.2.1 Comparison with other datasets

In this section, we describe how we trained AT-MAFCN by utilizing different building datasets, and evaluated the trained models on the test set of the WHU-Mix building dataset. The details of AT-MAFCN are described in Section 4.2.2. The compared building datasets were the Massachusetts buildings dataset (Mnih, 2013), the Inria dataset (Maggiori et al., 2017) (with Kitsap County excluded), the CrowdAI dataset (Mohanty, 2018), the WHU building dataset (Ji et al., 2019), and the newly built WHU-Mix building dataset (trainval set). As the geographical regions of the WHU-Mix test set are far from the regions of all of the above-mentioned datasets, the results obtained can be used to fairly measure the ability of the various datasets to handle out-of-distribution data.

The results are presented in Table 3. The highest mIoU is achieved with the WHU-Mix building dataset (trainval set), which can be attributed to the diversity and large scale of the WHU-Mix building dataset, enabling the model to learn domain-invariant and robust features that facilitate the extraction of buildings that have not been seen before. The Massachusetts buildings dataset gives the most unsatisfactory result, which means that a dataset of both a very small size and low diversity cannot meet the requirement of a strong generalization capability. The Inria dataset (with Kitsap County excluded) and the WHU building dataset yield mIoU scores that are 6–8% lower than that for the WHU-Mix building dataset (trainval set), due to the relatively small scale and low diversity of these two datasets. Moreover, the label quality of the Inria dataset is unsatisfactory, which affects the performance of the building extraction model. Surprisingly, although the size of the CrowdAI dataset far exceeds that of the WHU-Mix building dataset, the performance with this dataset is very poor, with a 14.68% mIoU gap to the WHU-Mix building dataset. We believe that the main reason for this is that the images in the CrowdAI dataset lack variation, and the buildings lack intra-class diversity, which was verified in our visual check, which results in the features learned by the model being domain-specific. From the above results, it can be seen that both the diversity and quantity of data have a great impact on the generalization performance of the building extraction model, and the diversity can play a more important role, because a dataset with a very large quantity but low diversity will still lead to a weak generalization capacity.

When taking a closer look at the IoU score for each city, it can be found that, even the best-performing dataset, i.e., the WHU-Mix building dataset, achieves very low IoU scores with the challenging DE_Potsdam and SD_Khartoum data. These results show the difficulty of practical building extraction, i.e., the data distributions of some test images are very far from the sample space covered by the training datasets. Therefore, there is a great need to develop domain generalization methods to further improve the generalization ability of building extraction models.

Table 3 Quantitative results of AT-MAFCN trained with the different building datasets on the test set of the WHU-Mix building dataset. The best result is highlighted in bold.

| Dataset | Number of image patches | mIoU (%) | IoU (%) | | | | | Training time (h) |
|---|---|---|---|---|---|---|---|---|
| | | | US_Kitsap | CN_Wuxi | NZ_Dunedin | DE_Potsdam | SD_Khartoum | |
| Massachusetts | 1359 | 22.37 | 28.39 | 9.84 | 46.42 | 7.50 | 19.97 | 2.1 |
| Inria | 14400 | 52.46 | 68.09 | 63.95 | 63.75 | 37.70 | 29.79 | 22.4 |
| CrowdAI | 341058 | 44.21 | 34.86 | 52.23 | 58.36 | **41.68** | 29.64 | 86.2 |
| WHU | 12430 | 51.33 | 59.75 | 61.35 | **84.53** | 20.06 | 35.07 | 10.7 |
| WHU-Mix (trainval set) | 43727 | **58.89** | **69.76** | **79.61** | 83.02 | 24.02 | **39.60** | 52.9 |

Fig. 7 shows 512 × 512 image patches from the five cities in the test set of the WHU-Mix building dataset. It can be seen that, in the US_Kitsap, CN_Wuxi, and NZ_Dunedin scenes, most of the datasets produce relatively accurate segmentation results. However, for the DE_Potsdam and SD_Khartoum scenes, the building extraction results are not as good. Overall, using the WHU-Mix building dataset (trainval set) results in the best performance, and produces the most complete building masks, especially in the red box areas of Fig. 7.

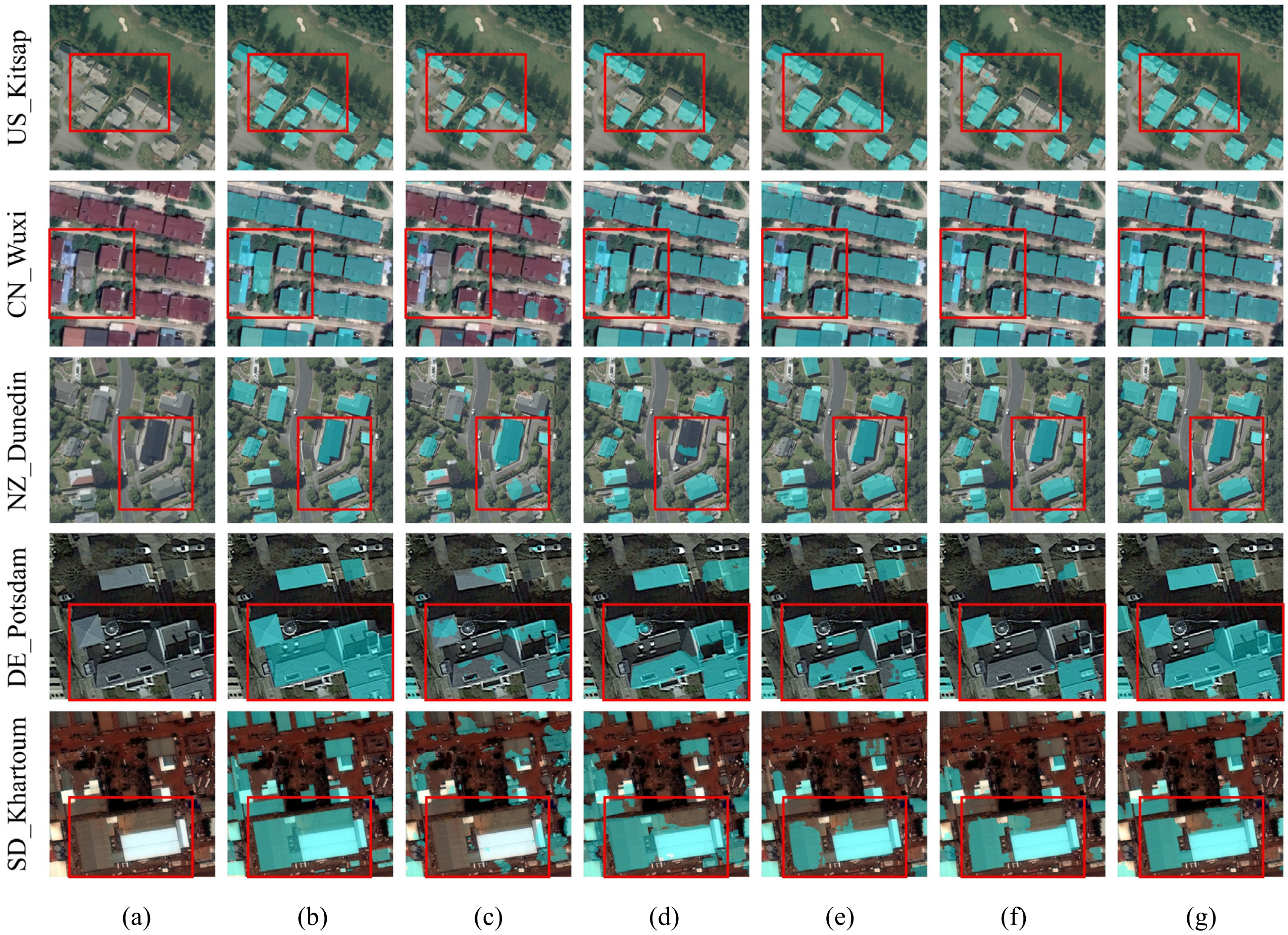

Fig. 7. Sample prediction results obtained by training AT-MAFCN with different building datasets for five 512 × 512 image patches. (a) Images. (b) Labels (cyan mask). (c)–(g) The results for the Massachusetts buildings dataset (Mnih, 2013), Inria dataset (Maggiori et al., 2017) (with Kitsap County excluded), CrowdAI dataset (Mohanty, 2018), WHU building dataset (Ji et al., 2019), and WHU-Mix building dataset (trainval set), respectively.

### 5.2.2 Influence of label quality

In this section, we investigate the influence of label quality on the building extraction performance. The experiment was conducted on three subsets of the Inria dataset—Western Tyrol, Vienna, and Kitsap County (Maggiori et al., 2017)—which are also included in the WHU-Mix building dataset. The original labels for Western Tyrol, Vienna, and Kitsap County have some significant errors and are of a poor quality (Fig. 3). The labeling errors were corrected, one by one, manually and carefully, in the construction of the WHU-Mix building dataset. We selected the images of Western Tyrol and Vienna as the training set to train AT-MAFCN, and the Kitsap County images were used as the test set, resulting in four cases: 1) training and testing with the original training and test sets; 2) training with the original training set and testing with the revised test set; 3) training with the revised training set and testing with the original test set; and 4) training and testing with the revised training and test sets.

The results are listed in Table 4. Firstly, we compare the results for the first case and the fourth case. The IoU score for the first case (using the original low-quality training and test sets) is only 49.89%, and the score for the fourth case (using the revised sets) is as high as 69.33%. The IoU score gap between the two results is as large as 19.44%. By comparing the results for the second case and the fourth case, training with the revised training set improves the IoU score from 59.76% to 69.33%, indicating that training data with a low label quality introduce a lot of noisy information that prevents the model from learning the effective information. By comparing the results for the third and fourth cases, testing with the revised test set results in an improvement of 13.13% in IoU score over testing with the original test set, indicating that a test set with inaccurate labels cannot reflect the true ability of a method. In summary, we conclude that the label quality of both the training set and test set have a significant influence on the building extraction performance. Therefore, developing a dataset with high-quality labels for practical building extraction is very important and necessary, which was one of the aims of this work.

Table 4 IoU scores (%) obtained on the Kitsap County data (before and after revision) with the model trained on the Western Tyrol and Vienna data (before and after revision).

| Training | | Testing/Kitsap County | |
|---|---|---|---|
| | | Original | Revised |
| Tyrol + | Original | 49.89 | 59.76 |
| Vienna | Revised | 56.20 | 69.33 |

Fig. 8 gives two examples of 512 × 512 image patches from the Kitsap County test set. Fig. 8 (b) shows the original labels of the two image patches with obvious errors, and Fig. 8 (c) and (d) show the prediction maps obtained by training with the original and revised training sets, respectively. It can be observed by comparing Fig. 8 (b) and Fig. 8 (d) that the use of low-quality labels for the evaluation leads to a lower performance and a failure of the fair evaluation. The comparison of Fig. 8 (c) and Fig. 8 (d) indicates the importance of a high-quality training set for improving the building extraction model performance.

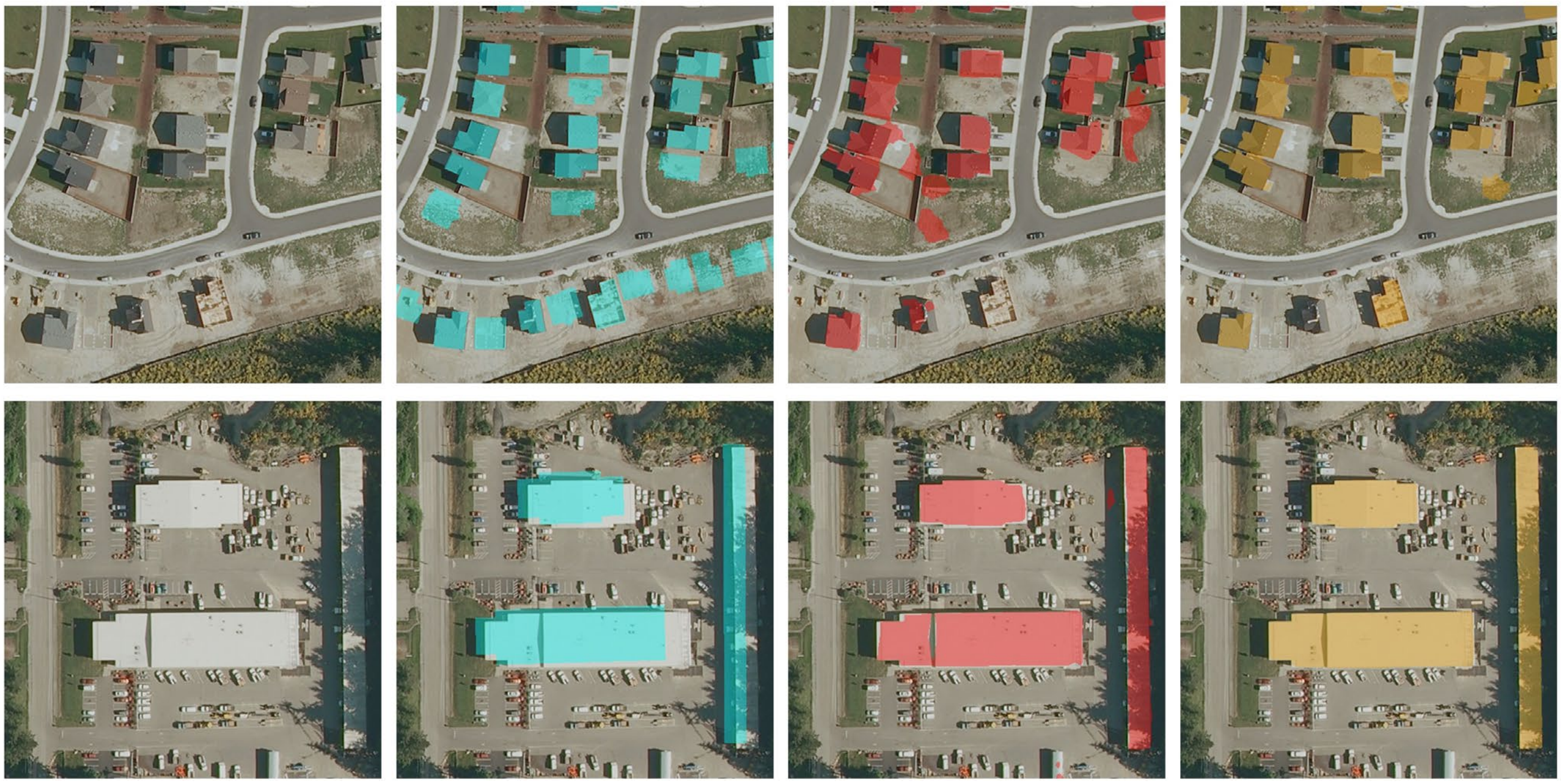

(a) (b) (c) (d)

Fig. 8. Two examples from the Kitsap County test set. (a) Images. (b) The original labels. (c) The predictions obtained by the model trained on the original training set. (d) The predictions obtained by the model trained on the revised training set.

## 5.3 Evaluation of the proposed domain generalization method

Section 5.3.1 compares the proposed domain generalization method, i.e., BSM, with the existing domain generalization methods. Section 5.3.2 describes the ablation study conducted for BSM, in which we evaluated the functions of the three submodules. Section 5.3.3 focuses on the universal adaptability of the proposed BSM method in different building extraction models.

### 5.3.1 Comparison with other domain generalization methods

In this section, we compare the generalization ability of the proposed BSM method and the recent domain generalization methods. AT-MAFCN was used as the classification model, and the methods were trained on the trainval set of the WHU-Mix building dataset, and tested in the test set of the WHU-Mix building dataset. The compared methods were a recently proposed image-level method known as Mixstyle (Zhou et al., 2021b), three feature-level methods, i.e., IBN-Net (Pan et al., 2018), MMLD (Matsuura and Harada, 2020), and LDDG (H. Li et al., 2020), and an image/feature combined method known as RobustNet (Choi et al., 2021). The compared methods, as well as the proposed method, are the few domain generalization methods that can be used in the absence of domain labels. A description of the compared methods has been given in Section 2.2. The image-level methods are basically independent of the classification model. However, the feature-level and image/feature combined methods required us to revise AT-MAFCN or add new losses in the training process. Taking IBN-Net as an example, we placed an instance normalization layer and a rectified linear unit (ReLU) layer at the end of the first three scales of the encoder of AT-MAFCN, followed by the IBN-b structure described in Pan et al. (2018).

The experimental results of the different methods are listed in Table 5. The result of AT-MAFCN trained without any domain generalization method is also listed as the baseline. The proposed BSM method obtains an mIoU score of as high as 72.06%, and shows a significant improvement (13.17%) over the baseline, representing a great success in improving the model's generalization ability. BSM also outperforms all the compared methods with respect to the mIoU score. Among the three feature-level methods, only IBN-Net improves the mIoU score of the baseline slightly, while MMLD and LDDG make the results even worse. The image-level Mixstyle method shows a slight improvement over the baseline. The image/feature combined method—RobustNet—is the second-best method; however, there is a 4.59% mIoU gap between RobustNet and the proposed BSM. We take the above results as an indication that image-level generalization is more stable than feature-level generalization, and can be used as a prerequisite for the feature-level generalization, to make it more robust. However, feature-level generalization may not be necessary as the proposed BSM method beats the very recent RobustNet by a large margin, demonstrating that a well-designed image-level method can be more powerful than a complex image/feature combined method.

At the individual city scale, it can be seen that the proposed BSM method improves the results of the baseline greatly in the most challenging DE_Potsdam (34% IoU improvement) and SD_Khartoum (9% IoU) scenarios, while the other methods are far less effective. RobustNet, in particular, even results in a 9% IoU decrease with the SD_Khartoum data. For the US_Kitsap, CN_Wuxi, and NZ_Dunedin data, the improvements of the proposed BSM method are relatively modest, and are at the same level as RobustNet. This can be explained by the fact that a higher improvement can be accessed when the room for improvement is larger. Nevertheless, the results for the DE_Potsdam and SD_Khartoum data indicate that the BSM method is more robust and shows a significant advantage over the other methods.

The training of the proposed method consumes 18 h more than the baseline, due to the data augmentation steps. However, it has a comparable efficiency with the compared methods. Considering the significant improvement of the proposed BSM method over the other five methods and the baseline, its efficiency can be considered to be satisfactory.

A visual comparison of the results of the different domain generalization methods is shown in Fig. 9. It can be seen that the proposed BSM method generates more accurate building masks than the other methods, especially with the DE_Potsdam and SD_Khartoum data.

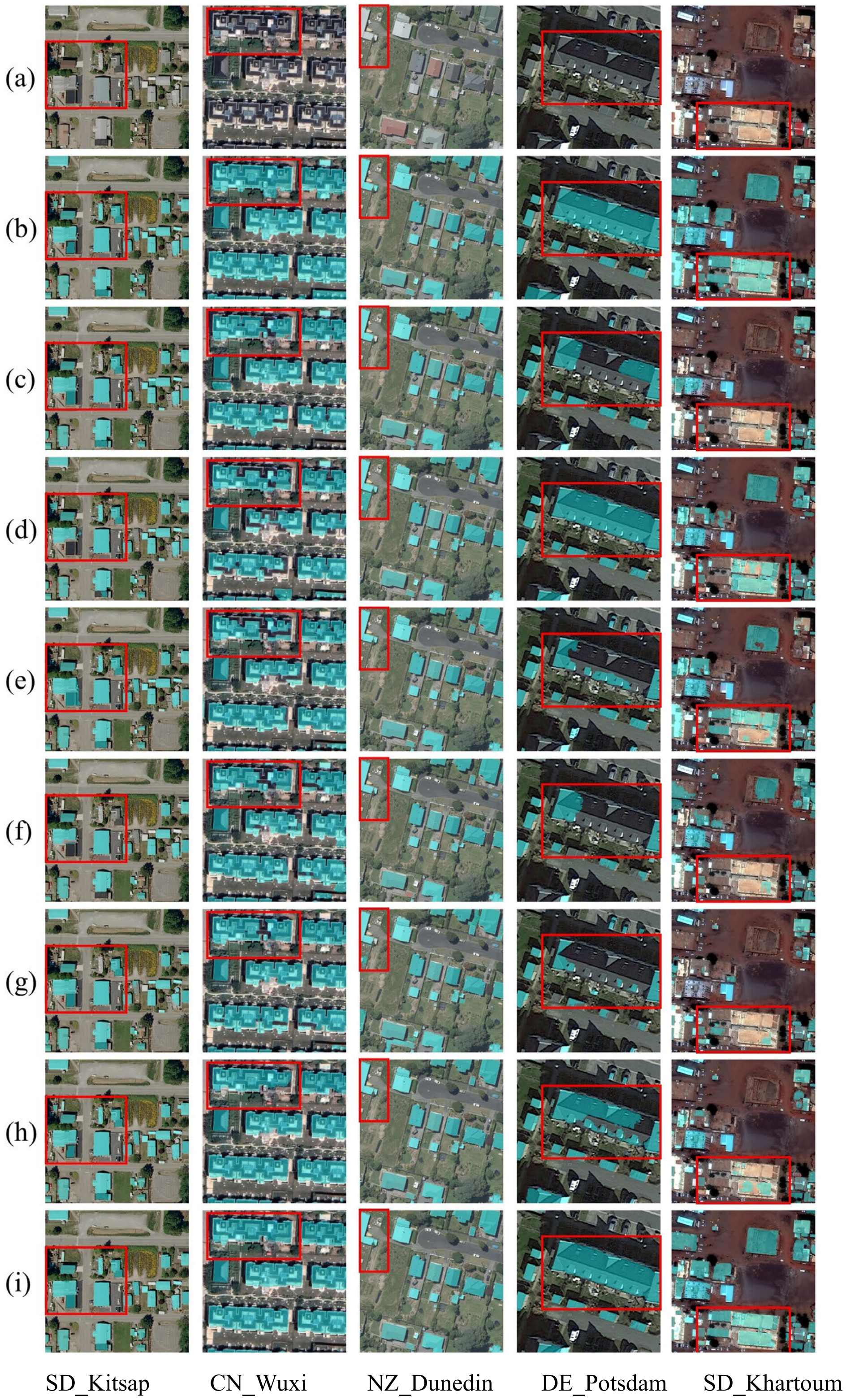

Fig. 9. Prediction results of the different domain generalization methods on five 512 × 512 image patches. (a) Images. (b) Labels (cyan mask). (c)–(i) The results of AT-MAFCN, Mixstyle (Zhou et al., 2021b), IBN-Net (Pan et al., 2018), MMLD (Matsuura and Harada, 2020), LDDG (H. Li et al., 2020), RobustNet (Choi et al., 2021), and the proposed BSM method, respectively.

Table 5 Quantitative results of the different domain generalization methods on the test set of the WHU-Mix building dataset. The best result is highlighted in bold.

| Method | mIoU (%) | IoU (%) | | | | | Training time (h) |
|---|---|---|---|---|---|---|---|
| | | US_Kitsap | CN_Wuxi | NZ_Dunedin | DE_Potsdam | SD_Khartoum | |
| AT-MAFCN (Baseline) | 58.89 | 69.76 | 79.61 | 83.02 | 24.02 | 39.60 | 52.9 |
| Mixstyle | 60.86 | 76.45 | 80.32 | 84.91 | 25.17 | 40.35 | 56.9 |
| IBN-Net | 59.91 | 75.77 | 79.29 | 85.96 | 21.13 | 41.97 | 69.0 |
| MMLD | 56.77 | 68.06 | 74.10 | 82.02 | 22.64 | 39.94 | 71.0 |
| LDDG | 58.02 | 71.21 | 79.92 | 81.25 | 26.53 | 31.37 | 55.2 |
| RobustNet | 67.47 | **81.16** | **84.92** | **88.98** | 47.48 | 33.30 | 74.9 |
| BSM (proposed) | **72.06** | 80.93 | 84.04 | 88.62 | **58.16** | **48.94** | 70.5 |

### 5.3.2 Ablation study

To further verify the effectiveness of the different parts of the proposed BSM domain generalization method, we investigated the performance gain with the use of the different submodules, i.e., geometric augmentation (GA), color augmentation (CA), style mixing (SM), and their combinations. The results are listed in Table 6. "AT-MAFCN (Baseline)" means training AT-MAFCN without BSM. By comparing the "GA", "CA", and "SM" submodules, it can be seen that GA achieves the best result. This is because buildings are highly variable in both shape and size, and only GA can handle the variations in sizes and shapes. For CA and SM, both of which help the model adapt to the variations of image colors and styles, SM shows a better performance than CA, showing its superiority in diversifying the image styles. It is worth noting that SM is the only submodule that improves the performance of AT-MAFCN on the SD_Khartoum data, while the other two submodules have a negative effect. The combinations of two submodules show better performances than the individual submodules, which shows that the three submodules are complementary and mutually reinforcing. Not unexpectedly, the best performance is achieved by the joint combination of the three submodules, verifying the effectiveness of the proposed BSM domain generalization method.

Compared with GA, CA, and their combinations, SM and the combinations including SM take a slightly longer time for the training; however, their efficiency is at the same level.

Table 6 Ablation study for the different submodules and their combinations in the proposed BSM method on the test set of the WHU-Mix building dataset. The best result is highlighted in bold.

| Submodule | mIoU (%) | IoU (%) | | | | | Training time (h) |
|---|---|---|---|---|---|---|---|
| | | US_Kitsap | CN_Wuxi | NZ_Dunedin | DE_Potsdam | SD_Khartoum | |
| AT-MAFCN (Baseline) | 58.89 | 69.76 | 79.61 | 83.02 | 24.02 | 39.60 | 52.9 |
| GA | 66.45 | 76.08 | 84.16 | 88.01 | 46.84 | 35.13 | 67.5 |
| CA | 61.03 | 77.39 | 80.91 | 86.29 | 32.21 | 27.55 | 66.5 |
| SM | 64.72 | 78.30 | 81.53 | 84.17 | 40.59 | 40.36 | 70.4 |
| GA + CA | 68.59 | **81.48** | **84.90** | **89.06** | 54.33 | 29.28 | 67.2 |
| GA + SM | 68.02 | 79.57 | 83.56 | 88.12 | 51.36 | 36.94 | 71.1 |
| CA + SM | 66.66 | 78.26 | 82.21 | 86.38 | 49.15 | 37.49 | 58.7 |
| GA + CA + SM (BSM) | **72.06** | 80.93 | 84.04 | 88.62 | **58.16** | **48.94** | 70.5 |

### 5.3.3 Evaluation of the universal adaptability of the domain generalization method

In the experiments described above, we implemented AT-MAFCN as the baseline model for the domain generalization baseline to extract buildings. However, it is important to note that the proposed domain generalization method is not especially designed for AT-MAFCN, and is a plug-and-play data augmentation module, which can be coupled with arbitrary modern deep learning based building extraction models. We imbedded BSM into two popular semantic segmentation models, i.e., DeepLab V3+ (Chen et al., 2018) and UNet (Ronneberger et al., 2015), to further investigate the robustness and performance of BSM. Training with the trainval set of the WHU-Mix building dataset and testing with the test set, we obtained the results listed in Table 7. It can be seen that BSM significantly improves the performance of the models by about 12% in mIoU, demonstrating its ability to improve the generalization ability of the existing models. However, there is a performance gap between the two models and the more sophisticated AT-MAFCN (Table 5, 72.06% mIoU), indicating that we should obtain a better performance when we equip a better building extraction model in the future with BSM. Nevertheless, the performance difference between AT-MAFCN and DeepLab V3+ is only 2%, indicating that developing domain generalization is currently more important than developing different segmentation structures, as the latter have reached a peak after several years of extensive studies.

Table 7 Quantitative results of the proposed BSM domain generalization method using different building extraction models.

| Building extraction model | | mIoU (%) | IoU (%) | | | | | Training time (h) |
|---|---|---|---|---|---|---|---|---|
| | | | US_Kitsap | CN_Wuxi | NZ_Dunedin | DE_Potsdam | SD_Khartoum | |
| DeepLab V3+ | Baseline | 58.10 | 65.92 | 81.39 | 76.78 | 35.40 | 26.05 | 54.5 |
| | Proposed | 70.37 | 77.68 | 83.93 | 86.60 | 56.69 | 46.78 | 66.3 |
| UNet | Baseline | 51.65 | 64.36 | 69.22 | 73.41 | 16.42 | 39.24 | 54.9 |
| | Proposed | 63.99 | 75.69 | 77.04 | 85.92 | 44.25 | 40.98 | 58.6 |

From the comprehensive evaluation of the proposed BSM method in Section 5.3, it is concluded that BSM not only exhibits a superior performance and robustness, compared to the commonly used data augmentation strategies and the more recent sophisticated methods, but it also has the advantage of being a portable and plug-and-play module. This indicates that style mixing, which is mainly a binary operation, is a powerful and universal strategy, as effective as and supplementary to the widely used geometric and color augmentation, which is a unary operation, for enlarging the distribution space of training data, which has not been recognized by the remote sensing community. We believe that BSM could become a new domain generalization baseline, not only for building extraction, but also for most applications involved with remote sensing image processing.

# 6. Conclusion

In this paper, we have described how we built a new diverse, large-scale, and high-quality building dataset, the WHU-Mix building dataset, to promote building extraction research toward more practical applications. The comparison experiments conducted between different building datasets verified the advantage of the WHU-Mix building dataset in promoting the performance of a building extraction model and properly evaluating the generalization ability of a model, and also the importance of diversity and high quality.

We also proposed a domain generalization method named batch style mixing (BSM), which can significantly improve the generalization ability of a building extraction model in a convenient, efficient, and plug-and-play manner. The classic data augmentation methods and the novel style mixing in BSM complement each other, providing the building extraction model with a vast data distribution space to learn the domain-invariant information. The extensive experiments demonstrated the superiority and robustness of BSM over both image-level and feature-level domain generalization methods, and also its ease of use. We believe that BSM could serve as a new domain generalization baseline for remote sensing image processing, and will not be limited to building extraction in the future.

# Declaration of Interest

The authors declare that they have no conflicts of interest.

# Acknowledgement

This work was supported by the National Natural Science Foundation of China (grant No. 42171430) and the State Key Program of the National Natural Science Foundation of China (grant No. 42030102).